\documentclass{article} % For LaTeX2e
\usepackage{iclr2025_conference,times}

\usepackage{amsmath,amsfonts,bm}

\def\eqref#1{equation~\ref{#1}}
\def\1{\bm{1}}

\DeclareMathAlphabet{\mathsfit}{\encodingdefault}{\sfdefault}{m}{sl}
\SetMathAlphabet{\mathsfit}{bold}{\encodingdefault}{\sfdefault}{bx}{n}

\usepackage{hyperref}
\usepackage{url}
\usepackage{float}
\usepackage{placeins}
\usepackage{tikz, tcolorbox}
\usepackage{tablefootnote}

\title{Uncovering Latent Chain of Thought Vectors in Large Language Models}

\author{
  Jason Zhang \\
  Stanford University \\
  \texttt{jasonbz@stanford.edu}
  \And
  Scott Viteri \\
  Stanford University \\
  \texttt{sviteri@stanford.edu} \\
}

\iclrfinalcopy % Uncomment for camera-ready version, but NOT for submission.
\begin{document}

\maketitle

\begin{abstract}
In this work, we examine how targeted perturbations in the activation space of Language Models (LMs) can encode complex reasoning patterns. We inject steering vectors, derived from LM activations, into LMs during inference time and study whether these vectors can induce Chain-of-Thought (CoT) reasoning in LMs without the need for natural language prompting. We demonstrate this approach on Llama3 8B Instruct and Mistral 7B v0.2 Instruct and show that activation-space interventions achieve competitive, if not superior, performance compared to traditional CoT prompting across multiple reasoning benchmarks, including GSM8k, MMLU, AGI Eval, and ARC AI2. These findings suggest that neural network activations can encode reasoning patterns, offering a new application of activation space manipulation as a tool for tuning model behavior.
\end{abstract}

\section{Introduction \& Related Works}
Language Models (LMs) have seen rapid adoption in recent years \citep{Zhao2023ASO}, yet still struggle with reasoning and accuracy \citep{huang2023surveyhallucinationlargelanguage, mondorf2024accuracyevaluatingreasoningbehavior}.  In response, Chain of Thought (CoT) reasoning has emerged as an effective way to enable LMs to break down complex problems and catch errors  \citep{Wei2022ChainOT, nye2022show}. However, methods used to steer models towards CoT like prompt engineering \citep{Radford2019LanguageMA}, model fine-tuning \citep{devlin2019bertpretrainingdeepbidirectional}, inference time compute \citep{snell2024scaling}, and RLHF \citep{ziegler2020finetuninglanguagemodelshuman} can be limited in effectiveness, or require substantial compute \citep{Casper2023Open,chen2023longlora}. 

Recent work has demonstrated that neural network weights themselves may contain representations of learned behaviors and knowledge. Studies like ROME \citep{meng2022locating} show that targeted interventions in the weight space of LMs can modify specific model behavior while preserving general capabilities. Similarly, approaches like LoRA \citep{hu2021lora} have revealed that small, rank-decomposed weight updates can effectively adapt model behavior. These insights suggest that small weight space manipulations of LMs might be a more efficient means than traditional full-parameter fine tuning or direct prompting. 

Building on this intuition, the research field of representation learning has emerged, operating on the belief that the internal activations of an LM carry symbolic meaning \citep{azaria2023internalstatellmknows}. Applications of this approach for model steering first originated with Plug-and-Play LMs as outlined by \citep{dathathri2020plugplaylanguagemodels}, in which a classifier was trained and used to steer layer activations closer towards desired activation structures that successfully reduced toxicity and strengthened sentiment control in model responses.

This approach was later followed by \citet{Subramani2022Extracting}, who first formally used a learned ``steering vector'' to alter the activation space of a frozen LM. This approach yielded strong control over model generation, achieving near-perfect model reconstruction for target memorized sentences. Similar work was later done by \citet{turner2023activation} and \citet{liu2024incontextvectorsmakingcontext} who used natural language prompting instead of gradient based approaches to extract steering vectors for similar goals of sentiment control and toxicity reduction. 

This work builds upon existing methods of using steering vectors to manipulate the activation space of LMs, investigating whether perturbations to the activation stream can steer models towards CoT behavior. We outline the process of creating steering vectors and evaluate their effectiveness on a variety of reasoning benchmarks (GSM8k, MMLU, ARC Challenge, AGI Eval) \citep{cobbe2021trainingverifierssolvemath, hendrycks2021measuringmassivemultitasklanguage, chollet2019measureintelligence, zhong2023agievalhumancentricbenchmarkevaluating}. Our experiments, conducted on Llama3 8B Instruct and Mistral 7B v0.2 Instruct \citep{dubey2024llama3herdmodels, jiang2023mistral7b}, demonstrate that this approach is not only effective but also competitive when compared to models using traditional CoT prompting techniques.

\begin{figure}[h]
    \centering
    \vspace{0.2cm}  % Reduce space above figure
    \includegraphics[width=0.8\columnwidth]{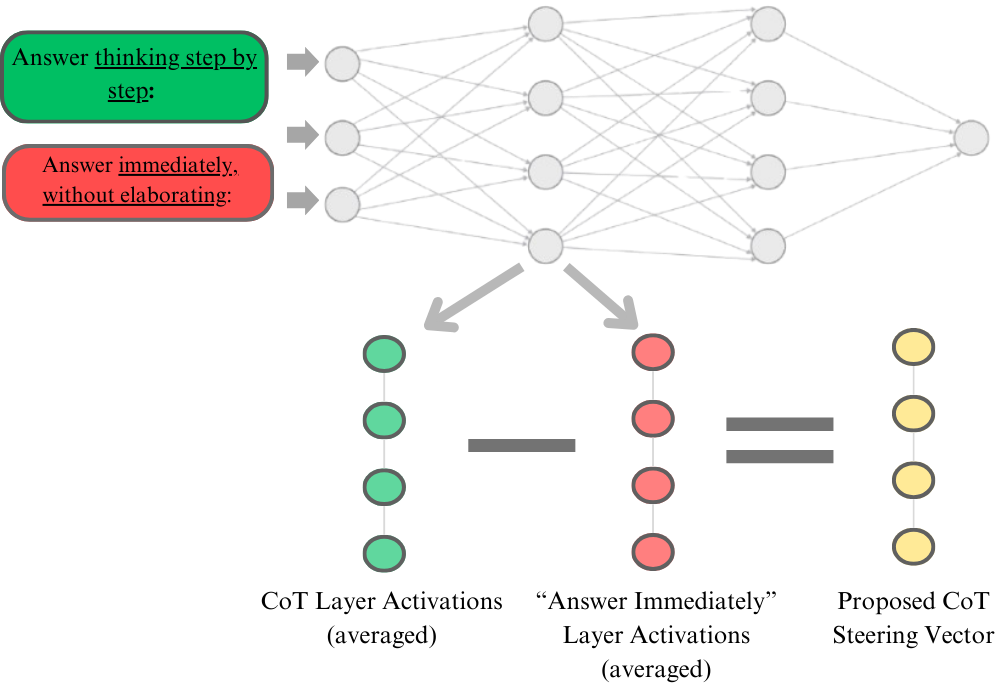}
    \caption{Overview of Steering Method Vector Creation Workflow -- we use two contrasting natural language prompts alongside questions from reasoning benchmarks to extract ``Think Step by Step'' and ``Answer Immediately'' layer residual stream activations from the LM. We then average these activations across token positions and across a corpus of reasoning questions, subtracting the two vectors to obtain a proposed ``CoT Steering Vector'' for injection.}
    
    \label{fig:method}
\end{figure}

\section{Methodology and Experimental Setup}
\subsection{Steering Vector Derivation}
To derive steering vectors from LM layer activations, we employ a contrastive approach inspired by \citet{panickssery2024steeringllama2contrastive}. Let \(\mathcal{D}\) denote the data-generating distribution over questions. In our experiments, \(\mathcal{D}\) is instantiated by a corpus of roughly 300 logic questions drawn from established reasoning benchmarks, composed as follows:
\begin{itemize}
    \item \textbf{Big Bench Lite:} 3 questions from each category (72 total),
    \item \textbf{MMLU:} 2 questions from each category in the development set (125 total),
    \item \textbf{GSM8k:} 100 questions from the training set.
\end{itemize}

For each question \(q \sim \mathcal{D}\), we create two contrasting inputs by appending a specific prompt:
\begin{itemize}
    \item \textbf{CoT:} ``Answer the question thinking step by step...'' 
    \item \textbf{Direct:} ``Answer the question immediately, without any elaboration...'' 
\end{itemize}
We denote these modified inputs as \(q_{\text{CoT}}\) and \(q_{\text{direct}}\), respectively.

Consider a fixed transformer layer \(l\) in the LM, which comprises a self-attention sublayer and an MLP sublayer, combined via residual connections. We extract activation vectors from the residual stream. In particular, let $\mathbf{a}^{(l)}_t(q_p)$
denote the activation vector at token position \(t\) in layer \(l\) when processing input \(q_p\) (with \(p \in \{\text{CoT}, \text{direct}\}\)); these activations depend on the generated text sampled from the LM's distribution \(P(\cdot \mid q_p)\).

We define the steering vector for a given question \(q\) as the difference between the expected activation vectors over token positions:
\[
\mathbf{v}(q) = \mathbb{E}_t \big[\mathbf{a}^{(l)}_t(q_{\text{CoT}})\big] - \mathbb{E}_t \big[\mathbf{a}^{(l)}_t(q_{\text{direct}})\big] \,.
\]
Finally, we aggregate over questions from \(\mathcal{D}\) to obtain the overall steering vector:
\[
\mathbf{v} = \mathbb{E}_{q \sim \mathcal{D}} \big[\mathbf{v}(q)\big] \,.
\]

This aggregated vector, once scaled by a coefficient \(c\), is used to steer the LM's output toward CoT reasoning. In the next section, we describe the specific injection strategies we employ.

\subsection{Steering Vector Injection}

To steer the LM, we inject the aggregated steering vector \(\mathbf{v}\) into the residual stream of layer \(l\), scaling it by a coefficient \(c\). This is accomplished using a PyTorch hook that modifies the output of the transformer layer (i.e., after its self-attention and feed-forward sublayers, including residual connection and layer normalization). In our experiments, this injection perturbs the hidden states passed to subsequent layers, thereby influencing the generated text.

We explore two injection strategies:
\begin{enumerate}
    \item \textbf{Continuous Injection:} The steering vector is injected at every token generation step with a low coefficient \(c\), following the approach of \citet{panickssery2024steeringllama2contrastive}.
    \item \textbf{Single Injection:} The steering vector is injected once during the processing of the input prompt.
\end{enumerate}

Through grid search, we determine optimal hyperparameters for both strategies. For example, with Llama3 8B Instruct we use \((l=16, c=20)\) for single injection and \((l=13, c=1)\) for continuous injection. For Mistral 7B v0.2 Instruct, we use \((l=15, c=10)\) and \((l=10, c=0.5)\) for the single and continuous strategies, respectively. The lower coefficients for continuous injection reflect the need to avoid over-steering during repeated applications at each generation step.

\section{Results}
\begin{table}[h]
\renewcommand{\arraystretch}{1.3}  % Increases vertical spacing between rows
\setlength{\tabcolsep}{8pt}        % Increases horizontal padding
\centering
\begin{tabular}{|p{4.0cm}|c|c|c|c|c|}
\hline
\rule{0pt}{3ex}\textbf{Model} & \textbf{GSM8K} & \textbf{MMLU} & \textbf{ARC-AI2} & \textbf{AGI-Eval}\tablefootnote{SAT Math Section} & \textbf{Average} \rule[-1.2ex]{0pt}{0pt}\\
\hline
\rule{0pt}{3ex}Llama3 8B Instruct:\newline \textbf{CoT Prompted} & 73.90 & \textbf{65.60} & 80.46 & 59.09 & 69.7625 \\
\hline
\rule{0pt}{3ex}Llama3 8B Instruct:\newline \textbf{Single Injection} & \textbf{79.15} & 64.20 & \textbf{81.23} & 61.40 & 71.495 \\
\hline 
\rule{0pt}{3ex}Llama3 8B Instruct:\newline \textbf{Continuous Injection} & 78.32 & 64.50 & 80.46 & \textbf{62.72} & \textbf{71.50} \\
\hline
\hline
\rule{0pt}{3ex}Mistral 7B v0.2 Instruct:\newline \textbf{CoT Prompted} & 50.72 & 48.95 & 60.75 & 40.00 & 50.105 \\
\hline 
\rule{0pt}{3ex}Mistral 7B v0.2 Instruct:\newline \textbf{Single Injection} & \textbf{51.40} & 48.20 & \textbf{66.64} & 38.70 & 51.235 \\
\hline
\rule{0pt}{3ex}Mistral 7B v0.2 Instruct:\newline \textbf{Continuous Injection} & 48.20 & \textbf{52.30} & 62.70 & \textbf{42.30} & \textbf{51.375} \\
\hline
\end{tabular}
\\\vspace{1mm}
%\footnotesize{*SAT Math Section}
\caption{Comparison of Steered Model Performance vs. CoT Prompted Model Performance. Bolded numbers denote the best scores observed across the 3 steering methods for each model.}
\label{tab:model-comparison}
\end{table}

Our experiments show that the proposed approach is competitive with CoT prompting, as shown in \ref{tab:model-comparison}. We find that steered systems outperform CoT prompted systems for 3 out of the 4 benchmarks for Llama3 8B, and all 4 benchmarks for Mistral 7B v0.2. Moreover, the approach demonstrates strong performance on AGI-Eval's SAT Math section, a benchmark whose data was not present in the steering vector data distribution, indicating generalization of reasoning performance.

We also find that the chosen method of steering vector injection (Continuous vs. Single Injection) can yield notable variation in performance. Specifically, in Mistral 7B v0.2 Instruct, we observe higher performance on ARC-AI2 (66.64\%) with Single Injection compared to Continuous Injection (62.70\%), while Continuous Injection performs better on MMLU compared to Single Injection (52.30\% vs. 48.20\%). These variations in performance across different steering methods show that the choice of injection strategy may be an important consideration for optimizing steered model performance. On average across the datasets, Continuous Injection yields slightly higher performance than Single Injection, which in turn performs slightly better than CoT Prompting.
 
\begin{figure}[h]
    \vspace{-0.1cm}  % Reduce space above figure
    \includegraphics[width=0.9\columnwidth]{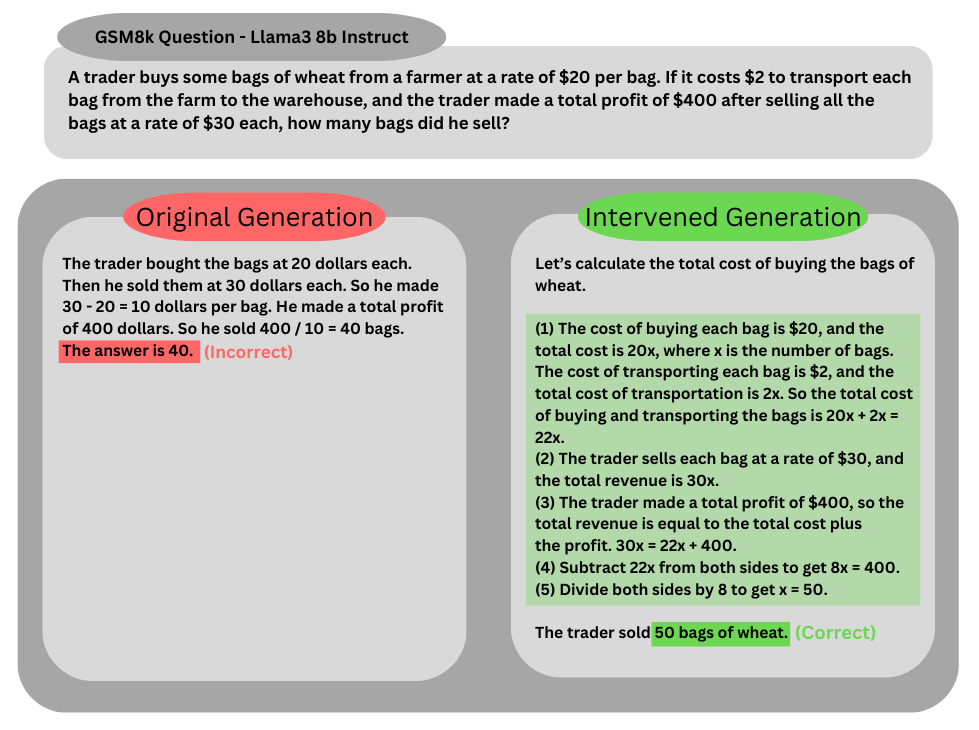}
    \vspace{-0.5cm}  
    \caption{Llama3 8B Instruct: Steered vs. Unsteered Response}
    \label{fig:qualitative_ex}
    \vspace{-0.2cm}  % Reduce space below figure
\end{figure}

We also observe a qualitative steering towards more structured, step-by-step reasoning outputs in steered systems, as shown in Figure~\ref{fig:qualitative_ex} and Appendix~\ref{app:a}, showcasing that steered models are able to retain natural language fluency during generation.

\section{Conclusion}
In this work, we demonstrated that steering vector injection can effectively induce CoT reasoning in LMs without relying on traditional prompting techniques, thereby suggesting that reasoning capabilities may be encoded into the activation and weight space of LMs. Our results across multiple benchmarks show this approach can match or exceed the performance of CoT prompting, and we note that changing the timing of injection during inference can result in both positive and negative fluctuations in performance. Future work in this direction might explore different methods for injecting these steering vectors at inference time to optimize performance and reliability, or further investigate the generalization capabilities of these steering vectors across different models, perhaps at a larger scale. 

\clearpage
\bibliography{./iclr2025_conference}

\begin{thebibliography}{25}
\providecommand{\natexlab}[1]{#1}
\providecommand{\url}[1]{\texttt{#1}}
\expandafter\ifx\csname urlstyle\endcsname\relax
  \providecommand{\doi}[1]{doi: #1}\else
  \providecommand{\doi}{doi: \begingroup \urlstyle{rm}\Url}\fi

\bibitem[Azaria \& Mitchell(2023)Azaria and Mitchell]{azaria2023internalstatellmknows}
Amos Azaria and Tom Mitchell.
\newblock The internal state of an llm knows when it’s lying.
\newblock In \emph{Findings of the Association for Computational Linguistics: EMNLP 2023}, pp.\  967--976, 2023.

\bibitem[Casper et~al.(2023)Casper, Davies, Shi, Krendl~Gilbert, Scheurer, Rando~Ramirez, Freedman, Korbak, Lindner, Freire, et~al.]{Casper2023Open}
Stephen Casper, Xander Davies, Claudia Shi, Thomas Krendl~Gilbert, J{\'e}r{\'e}my Scheurer, Javier Rando~Ramirez, Rachel Freedman, Tomasz Korbak, David Lindner, Pedro Freire, et~al.
\newblock Open problems and fundamental limitations of reinforcement learning from human feedback.
\newblock \emph{Transactions on Machine Learning Research}, 2023.

\bibitem[Chen et~al.(2024)Chen, Qian, Tang, Lai, Liu, Han, and Jia]{chen2023longlora}
Yukang Chen, Shengju Qian, Haotian Tang, Xin Lai, Zhijian Liu, Song Han, and Jiaya Jia.
\newblock Longlora: Efficient fine-tuning of long-context large language models.
\newblock \emph{ICLR}, 2024.

\bibitem[Chollet(2019)]{chollet2019measureintelligence}
François Chollet.
\newblock On the measure of intelligence, 2019.
\newblock URL \url{https://arxiv.org/abs/1911.01547}.

\bibitem[Cobbe et~al.(2021)Cobbe, Kosaraju, Bavarian, Chen, Jun, Kaiser, Plappert, Tworek, Hilton, Nakano, Hesse, and Schulman]{cobbe2021trainingverifierssolvemath}
Karl Cobbe, Vineet Kosaraju, Mohammad Bavarian, Mark Chen, Heewoo Jun, Lukasz Kaiser, Matthias Plappert, Jerry Tworek, Jacob Hilton, Reiichiro Nakano, Christopher Hesse, and John Schulman.
\newblock Training verifiers to solve math word problems, 2021.
\newblock URL \url{https://arxiv.org/abs/2110.14168}.

\bibitem[Dathathri et~al.(2020)Dathathri, Madotto, Lan, Hung, Frank, Molino, Yosinski, and Liu]{dathathri2020plugplaylanguagemodels}
Sumanth Dathathri, Andrea Madotto, Janice Lan, Jane Hung, Eric Frank, Piero Molino, Jason Yosinski, and Rosanne Liu.
\newblock Plug and play language models: A simple approach to controlled text generation, 2020.
\newblock URL \url{https://arxiv.org/abs/1912.02164}.

\bibitem[Devlin et~al.(2019)Devlin, Chang, Lee, and Toutanova]{devlin2019bertpretrainingdeepbidirectional}
Jacob Devlin, Ming-Wei Chang, Kenton Lee, and Kristina Toutanova.
\newblock {BERT}: Pre-training of deep bidirectional transformers for language understanding.
\newblock In Jill Burstein, Christy Doran, and Thamar Solorio (eds.), \emph{Proceedings of the 2019 Conference of the North {A}merican Chapter of the Association for Computational Linguistics: Human Language Technologies, Volume 1 (Long and Short Papers)}, pp.\  4171--4186, Minneapolis, Minnesota, June 2019. Association for Computational Linguistics.
\newblock \doi{10.18653/v1/N19-1423}.
\newblock URL \url{https://aclanthology.org/N19-1423}.

\bibitem[Dubey et~al.(2024)Dubey, Jauhri, Pandey, Kadian, Al-Dahle, Letman, Mathur, Schelten, Yang, Fan, et~al.]{dubey2024llama3herdmodels}
Abhimanyu Dubey, Abhinav Jauhri, Abhinav Pandey, Abhishek Kadian, Ahmad Al-Dahle, Aiesha Letman, Akhil Mathur, Alan Schelten, Amy Yang, Angela Fan, et~al.
\newblock The llama 3 herd of models.
\newblock \emph{arXiv. org}, 2024.

\bibitem[Hendrycks et~al.(2021)Hendrycks, Burns, Basart, Zou, Mazeika, Song, and Steinhardt]{hendrycks2021measuringmassivemultitasklanguage}
Dan Hendrycks, Collin Burns, Steven Basart, Andy Zou, Mantas Mazeika, Dawn Song, and Jacob Steinhardt.
\newblock Measuring massive multitask language understanding.
\newblock In \emph{International Conference on Learning Representations}, 2021.

\bibitem[Hu et~al.(2021)Hu, Shen, Wallis, Allen-Zhu, Li, Wang, Wang, and Chen]{hu2021lora}
Edward~J Hu, Yelong Shen, Phillip Wallis, Zeyuan Allen-Zhu, Yuanzhi Li, Shean Wang, Lu~Wang, and Weizhu Chen.
\newblock Lora: Low-rank adaptation of large language models.
\newblock \emph{arXiv preprint arXiv:2106.09685}, 2021.

\bibitem[Huang et~al.(2024)Huang, Yu, Ma, Zhong, Feng, Wang, Chen, Peng, Feng, Qin, and Liu]{huang2023surveyhallucinationlargelanguage}
Lei Huang, Weijiang Yu, Weitao Ma, Weihong Zhong, Zhangyin Feng, Haotian Wang, Qianglong Chen, Weihua Peng, Xiaocheng Feng, Bing Qin, and Ting Liu.
\newblock A survey on hallucination in large language models: Principles, taxonomy, challenges, and open questions.
\newblock \emph{ACM Trans. Inf. Syst.}, November 2024.
\newblock ISSN 1046-8188.
\newblock \doi{10.1145/3703155}.
\newblock URL \url{https://doi.org/10.1145/3703155}.
\newblock Just Accepted.

\bibitem[Jiang et~al.(2023)Jiang, Sablayrolles, Mensch, Bamford, Chaplot, de~las Casas, Bressand, Lengyel, Lample, Saulnier, Lavaud, Lachaux, Stock, Scao, Lavril, Wang, Lacroix, and Sayed]{jiang2023mistral7b}
Albert~Q. Jiang, Alexandre Sablayrolles, Arthur Mensch, Chris Bamford, Devendra~Singh Chaplot, Diego de~las Casas, Florian Bressand, Gianna Lengyel, Guillaume Lample, Lucile Saulnier, Lélio~Renard Lavaud, Marie-Anne Lachaux, Pierre Stock, Teven~Le Scao, Thibaut Lavril, Thomas Wang, Timothée Lacroix, and William~El Sayed.
\newblock Mistral 7b, 2023.
\newblock URL \url{https://arxiv.org/abs/2310.06825}.

\bibitem[Liu et~al.(2024)Liu, Ye, Xing, and Zou]{liu2024incontextvectorsmakingcontext}
Sheng Liu, Haotian Ye, Lei Xing, and James~Y Zou.
\newblock In-context vectors: Making in context learning more effective and controllable through latent space steering.
\newblock In \emph{Forty-first International Conference on Machine Learning}, 2024.

\bibitem[Meng et~al.(2022)Meng, Bau, Andonian, and Belinkov]{meng2022locating}
Kevin Meng, David Bau, Alex Andonian, and Yonatan Belinkov.
\newblock Locating and editing factual associations in gpt.
\newblock \emph{Advances in Neural Information Processing Systems}, 35:\penalty0 17359--17372, 2022.

\bibitem[Mondorf \& Plank(2024)Mondorf and Plank]{mondorf2024accuracyevaluatingreasoningbehavior}
Philipp Mondorf and Barbara Plank.
\newblock Beyond accuracy: Evaluating the reasoning behavior of large language models -- a survey, 2024.
\newblock URL \url{https://arxiv.org/abs/2404.01869}.

\bibitem[Nye et~al.(2022)Nye, Andreassen, Gur-Ari, Michalewski, Austin, Bieber, Dohan, Lewkowycz, Bosma, Luan, Sutton, and Odena]{nye2022show}
Maxwell Nye, Anders~Johan Andreassen, Guy Gur-Ari, Henryk Michalewski, Jacob Austin, David Bieber, David Dohan, Aitor Lewkowycz, Maarten Bosma, David Luan, Charles Sutton, and Augustus Odena.
\newblock Show your work: Scratchpads for intermediate computation with language models, 2022.
\newblock URL \url{https://openreview.net/forum?id=iedYJm92o0a}.

\bibitem[Panickssery et~al.(2024)Panickssery, Gabrieli, Schulz, Tong, Hubinger, and Turner]{panickssery2024steeringllama2contrastive}
Nina Panickssery, Nick Gabrieli, Julian Schulz, Meg Tong, Evan Hubinger, and Alexander~Matt Turner.
\newblock Steering llama 2 via contrastive activation addition, 2024.
\newblock URL \url{https://arxiv.org/abs/2312.06681}.

\bibitem[Radford et~al.(2019)Radford, Wu, Child, Luan, Amodei, Sutskever, et~al.]{Radford2019LanguageMA}
Alec Radford, Jeffrey Wu, Rewon Child, David Luan, Dario Amodei, Ilya Sutskever, et~al.
\newblock Language models are unsupervised multitask learners.
\newblock \emph{OpenAI blog}, 1\penalty0 (8):\penalty0 9, 2019.

\bibitem[Snell et~al.(2024)Snell, Lee, Xu, and Kumar]{snell2024scaling}
Charlie Snell, Jaehoon Lee, Kelvin Xu, and Aviral Kumar.
\newblock Scaling llm test-time compute optimally can be more effective than scaling model parameters.
\newblock \emph{arXiv e-prints}, pp.\  arXiv--2408, 2024.

\bibitem[Subramani et~al.(2022)Subramani, Suresh, and Peters]{Subramani2022Extracting}
Nishant Subramani, Nivedita Suresh, and Matthew~E Peters.
\newblock Extracting latent steering vectors from pretrained language models.
\newblock In \emph{Findings of the Association for Computational Linguistics: ACL 2022}, pp.\  566--581, 2022.

\bibitem[Turner et~al.(2023)Turner, Thiergart, Leech, Udell, Vazquez, Mini, and MacDiarmid]{turner2023activation}
Alexander~Matt Turner, Lisa Thiergart, Gavin Leech, David Udell, Juan~J Vazquez, Ulisse Mini, and Monte MacDiarmid.
\newblock Activation addition: Steering language models without optimization.
\newblock \emph{arXiv preprint arXiv:2308.10248}, 2023.

\bibitem[Wei et~al.(2024)Wei, Wang, Schuurmans, Bosma, Ichter, Xia, Chi, Le, and Zhou]{Wei2022ChainOT}
Jason Wei, Xuezhi Wang, Dale Schuurmans, Maarten Bosma, Brian Ichter, Fei Xia, Ed~H. Chi, Quoc~V. Le, and Denny Zhou.
\newblock Chain-of-thought prompting elicits reasoning in large language models.
\newblock In \emph{Proceedings of the 36th International Conference on Neural Information Processing Systems}, NIPS '22, Red Hook, NY, USA, 2024. Curran Associates Inc.
\newblock ISBN 9781713871088.

\bibitem[Zhao et~al.(2023)Zhao, Zhou, Li, Tang, Wang, Hou, Min, Zhang, Zhang, Dong, Du, Yang, Chen, Chen, Jiang, Ren, Li, Tang, Liu, Liu, Nie, and rong Wen]{Zhao2023ASO}
Wayne~Xin Zhao, Kun Zhou, Junyi Li, Tianyi Tang, Xiaolei Wang, Yupeng Hou, Yingqian Min, Beichen Zhang, Junjie Zhang, Zican Dong, Yifan Du, Chen Yang, Yushuo Chen, Z.~Chen, Jinhao Jiang, Ruiyang Ren, Yifan Li, Xinyu Tang, Zikang Liu, Peiyu Liu, Jianyun Nie, and Ji~rong Wen.
\newblock A survey of large language models.
\newblock \emph{ArXiv}, abs/2303.18223, 2023.
\newblock URL \url{https://api.semanticscholar.org/CorpusID:257900969}.

\bibitem[Zhong et~al.(2024)Zhong, Cui, Guo, Liang, Lu, Wang, Saied, Chen, and Duan]{zhong2023agievalhumancentricbenchmarkevaluating}
Wanjun Zhong, Ruixiang Cui, Yiduo Guo, Yaobo Liang, Shuai Lu, Yanlin Wang, Amin Saied, Weizhu Chen, and Nan Duan.
\newblock Agieval: A human-centric benchmark for evaluating foundation models.
\newblock In \emph{Findings of the Association for Computational Linguistics: NAACL 2024}, pp.\  2299--2314, 2024.

\bibitem[Ziegler et~al.(2020)Ziegler, Stiennon, Wu, Brown, Radford, Amodei, Christiano, and Irving]{ziegler2020finetuninglanguagemodelshuman}
Daniel~M. Ziegler, Nisan Stiennon, Jeffrey Wu, Tom~B. Brown, Alec Radford, Dario Amodei, Paul Christiano, and Geoffrey Irving.
\newblock Fine-tuning language models from human preferences, 2020.
\newblock URL \url{https://arxiv.org/abs/1909.08593}.

\end{thebibliography}
\bibliographystyle{iclr2025_conference}
\newpage
\appendix
\section{Appendix}

\subsection{Qualitative Examples of Steered Responses - Llama3 8B and Mistral}\label{app:a}
\begin{figure}[h]
    \vspace{-0.1cm}  % Reduce space above figure
    \includegraphics[width=\columnwidth]{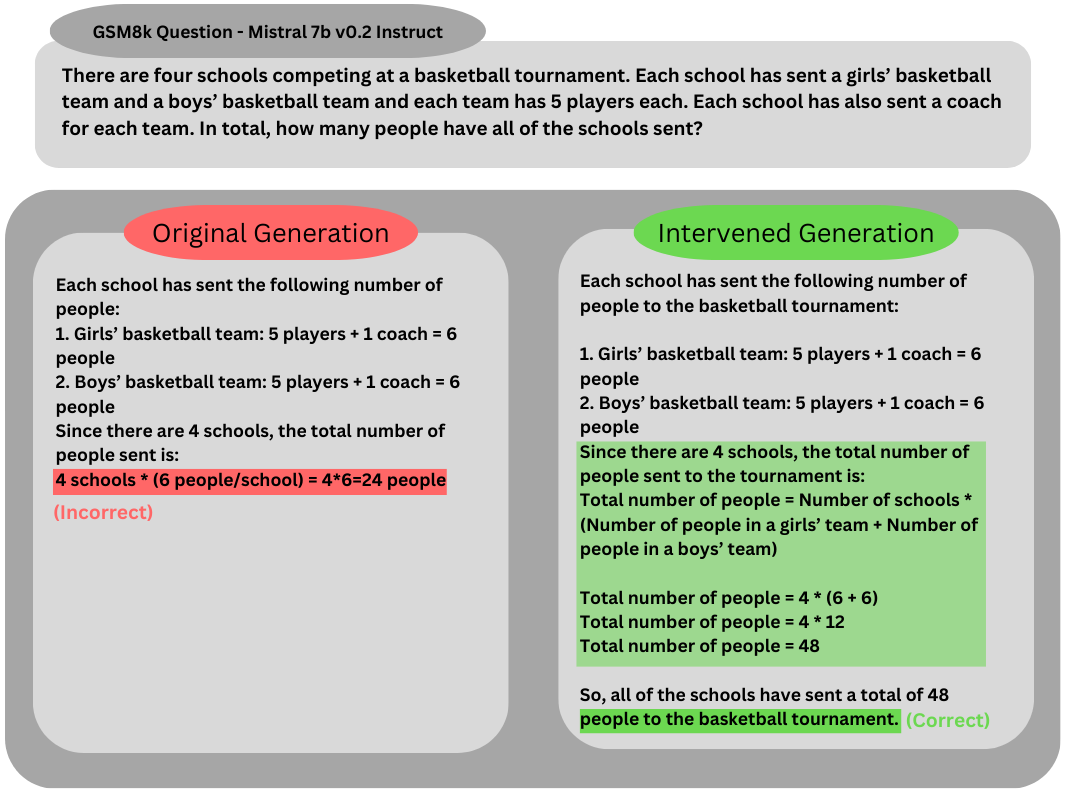}
    \vspace{-0.7cm}  % Reduce space between image and caption
    \caption{Mistral 7B v0.2 Instruct: Steered vs. Baseline Response}
\end{figure}

\end{document}